\documentclass[preprint,12pt]{elsarticle}

\usepackage{graphicx}
\usepackage{booktabs}
\usepackage{tabularx}

\usepackage{amssymb}
\usepackage{subfigure}
\usepackage{array}
\usepackage{lscape}
\usepackage{dcolumn,longtable,hhline,colortbl}
\usepackage[rotateright]{rotating}
\usepackage{multirow}
\usepackage[section] {placeins}
\usepackage{amsmath}
\usepackage{booktabs}
\usepackage{enumerate}
\usepackage{caption}
\usepackage{threeparttable}
\usepackage{url}
\usepackage{lscape}
\usepackage{algorithm}
\usepackage{algorithmic}

\newtheorem{example}{Example}[section]
\makeatletter

\begin{document}

\begin{frontmatter}

\title{A \emph{Physarum}-Inspired Approach to Optimal Supply Chain Network Design at Minimum Total Cost with Demand Satisfaction}

\author[address1]{Xiaoge Zhang}
\author[address4]{Andrew Adamatzky}
\author[address6]{Xin-She Yang}
\author[address5]{Hai Yang}
\author[address3]{Sankaran Mahadevan}
\author[address1,address3]{Yong Deng\corref{label1}}

\cortext[label1]{Corresponding author: Yong Deng, School of
Computer and Information Science, Southwest University, Chongqing,
400715, China. Email address: ydeng@swu.edu.cn,
prof.deng@hotmail.com. Tel/Fax:(86-023)68254555.}
\address[address1]{School of Computer and Information Science, Southwest University, Chongqing 400715, China}
\address[address4]{Unconventional Computing Center, University of the West of England, Bristol BS16 1QY, UK}
\address[address6]{School of Science and Technology, Middlesex University, London NW4 4BT, UK}
\address[address5]{Department of Civil and Environmental Engineering, The Hong Kong University of Science and Technology, Clear Water Bay, Kowloon, Hong Kong}
\address[address3]{School of Engineering, Vanderbilt University, Nashville, TN, 37235, USA}

\begin{abstract}
A supply chain is a system which moves products from a supplier to  customers. The supply chains are ubiquitous. They play a key role in all economic activities. Inspired by biological principles of nutrients' distribution in protoplasmic networks of slime mould \emph{Physarum polycephalum} we propose a novel algorithm for a supply chain design.
The algorithm handles the supply networks where capacity investments and product flows are variables. The networks
are constrained by a need to satisfy product demands. Two features of the slime mould are adopted in our algorithm.
The first is the continuity of a flux during the iterative process, which is used in real-time update of
the costs associated with the supply links. The second feature is adaptivity.  The supply chain can converge to
an equilibrium state when costs are changed.  Practicality and flexibility of our algorithm is illustrated on numerical examples.
\end{abstract}


\end{frontmatter}


\section{Introduction}
A supply chain is a network of suppliers, manufacturers, storage houses, and distribution centers organized
to acquire raw materials, convert these raw materials to finished products, and distribute these products to customers \cite{liu2013supply,zhang2014integrating,yu2013competitive}.
With the globalization of market economies, for many companies, especially for the high tech companies, such as Samsung, Apple, and IBM, their customers are all over the world and the components are also
distributed in many places ranging from Taiwan to South Africa. To design an efficient supply chain network the
enterprises must identify optimal capacities associated with various supply activities and the optimal production quantities, storage volumes as well as the shipments.They also must take into consideration the cost related with each activity. The cost, including the shipment, the shrinking resources of manufacturing bases, varies from day to day. From the practical standpoint, it is meaningful for us to consider these factors so that the sum of strategic, tactical, and operational costs can be minimized. In this way, the design of the supply chain network should be conducted in a rigorous way so that we can provide an insight into this problem from the system-wide view.

In the past decades the issue of designing the supply chain network got a great deal of attention, see e.g.
\cite{ma2006model,santoso2005stochastic,zhou2002balanced,trkman2009supply,altiparmak2009steady,ahmadi2010incorporating,nagurney2010supply,bilgen2010application}.
In 1998 Beamon \cite{beamon1998supply} presented an integrated supply chain network design model formulated as a multi-commodity mixed integer program and treated the capacity associated with each link as a known parameter. Two years later, Sabri and Beamon developed another approach to optimize the strategic and operational planning in the supply chains design problem using a multi-objective function. However, in their model, the cost associated with each link was a linear function and thus the model did nit capture the reality of dynamical networks, which are prone to congestions. Handfield and Nichols \cite{handfield2002supply} also employed discrete variables in the formulation of supply chain network model and this model was faced with the same problem mentioned above. Recently, Nagurney \cite{nagurney2010optimal} presented a framework for supply chain network design and redesign at minimal total cost subjecting to the demand satisfaction from a system-optimization perspective. They employed Lagrange multiplier to deal with the constraint associated with the link which made the model complicated with many variables.


Computer scientists and engineers are often looking into behaviour, mechanics, physiology of living systems to uncover novel principles of distributed sensing, information processing and decision making that could be adopted in development of future and emergent computing paradigms, architectures and implementations. One of the most popular nowadays
living computing substrates is a slime mould \emph{Physarum Polycephalum}.

Plasmodium is a vegetative stage of acellular slime mould \emph{P. polycephalum}, a single cell with many nuclei, which feeds on microscopic particles~\cite{stephenson_2000}. When foraging for its food the plasmodium propagates towards sources of food, surrounds them, secretes enzymes and digests the food; it may form a congregation of protoplasm covering the food source. When several sources of nutrients are scattered in the plasmodium's range, the plasmodium forms a network of protoplasmic tubes connecting the masses of protoplasm at the food sources.


In laboratory experiments and theoretical studies it is shown that the slime mould can solve many graph theoretical problems, including finding the shortest path \cite{nakagaki2000intelligence,zhang2013route,zhang2014physarum,tero2006physarum,zhang2013solving,zhang2013biologically},
connecting different arrays of food sources in an efficient manner \cite{nakagaki2007minimum,adamatzky2013route,tero2008flow}, network design \cite{tero2010rules,adamatzky2012bioevaluation,adamatzky2011approximating}.

\emph{Physarum} can be considered as a parallel computing substrate with distributed sensing, parallel information
processing and concurrent decision making. When the slime mould colonizes several sources of nutrients it dynamically updates thickness of its protoplasmic tubes, depending on how much nutrients is left in any particular sources and proximity of source of repellents, gradients of humidity and illumination~\cite{adamatzky_physarummachines,Andy2014}. This dynamical updating of the protoplasmic networks inspired us to employ principle of \emph{Physarum} foraging behaviour
to solve the supply chain network design problem aiming to minimize total costs of transportation and redistribution
of goods and services. In the  \emph{Physarum}-inspired algorithm proposed we consider links' capacities as design variables and use continuous functions to represent costs of the links. Based on the system-optimization technique developed for supply chain network integration \cite{nagurney2009system,nagurney2010multi}, we abstract the economic activities associated with a firm as a network. We make full use of two features of \emph{Physarum}:
a continuity of the flux during the iterative process and the protoplasmic network adaptivity, or reconfiguration.


The paper is structured as follows. In Section 2,  we introduce the supply chain network design model and our latest researches related to \emph{P. polycephalum}. In Section 3, we propose an approach to the supply chain network design problem based on \emph{Physarum} model. In Section 4, numerical examples are used to illustrate the flow of the proposed method and the methods's efficiency. We present our conclusions and provide suggestions for further studies  in Section 5.

\section{Preliminaries}

In this section, the supply chain network design model and the \emph{Physarum} model are introduced.

\subsection{The Supply Chain Network Design Model \cite{nagurney2010optimal}}

\begin{figure}[!ht]
\centering
\includegraphics[width=3in]{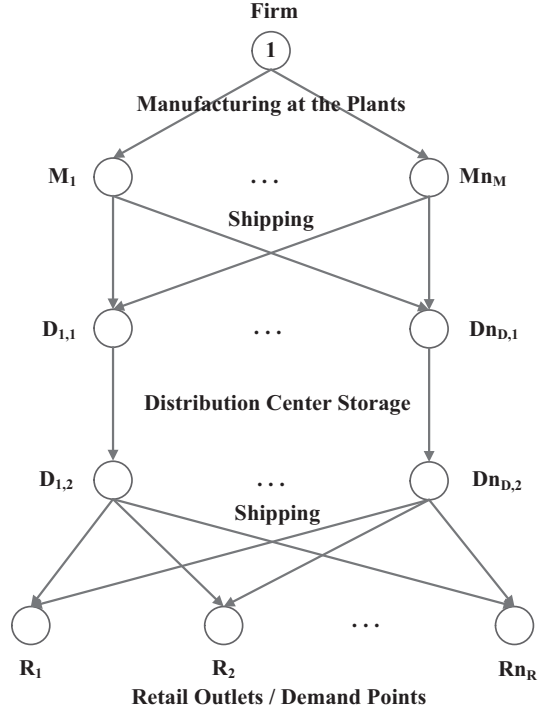}
\caption{The supply chain network topology}
\label{Topology}
\end{figure}

Consider the supply chain network shown in Fig. \ref{Topology}: a firm corresponding to node 1 aims at delivering the goods or products to the bottom level corresponding to the retail outlets. The links connecting the source node with the destination nodes represent the activities of production, storage and transportation of good or services.
Different network topologies corresponds to different supply chain network problems. In this paper, we assume that there exists only one path linking node 1 with each destination node, which can ensure that the demand at each retail outlet can be satisfied.

As shown in Fig. \ref{Topology},  the firm takes  into consideration $n_M$ manufacturers, $n_D$ distribution centers when $n_R$ retailers with demands  ${d_{{R_1}}},{d_{{R_2}}}, \cdots ,{d_{{R_{{n_R}}}}}$ must be served. The node $1$ in the first layer is linked with the possible $n_M$ manufacturers, which are represented as ${M_1},{M_2}, \cdots ,{M_{{n_M}}}$. These edges in the manufacturing level are associated with the possible distribution center nodes, which are expressed by ${D_{1,1}},{D_{2,1}}, \cdots ,{D_{{n_D},1}}$. These links mean the possible shipment between the manufacturers and the distribution centers. The links connecting ${D_{1,1}},{D_{2,1}}, \cdots ,{D_{{n_D},1}}$
with ${D_{1,2}},{D_{2,2}}, \cdots ,{D_{{n_D},2}}$ reflect the possible storage links. The links between ${D_{1,2}},{D_{2,2}}, \cdots ,{D_{{n_D},2}}$ and ${R_1},{R_1}, \cdots ,{R_{{n_R}}}$ denote the possible shipment links connecting the storage centers with the retail outlets.

Let a supply chain network be represented by a graph $G(N,L)$, where $N$ is a  set of nodes and $L$ is  a set of links. Each links in the network is associated with a cost function and the cost reflects the total cost of all the specific activities in the supply chain network, such as the transport of the product, the delivery of the product, etc. The cost related with link $a$ is expressed by ${\widehat c_a}$. A path $p$ connecting node 1 with a retail node shown in Fig. \ref{Topology} denotes the whole activities related with manufacturing the products, storing them and transporting them, etc. Assume $w_k$ denotes the set of source and destination nodes $(1, R_k)$ and $P_{w_k}$ represents the set of alternative associated possible supply chain network processes joining $(1,R_k)$. Then $P$ means the set of all paths joining $(1,R_k)$ while $x_p$ denotes the flow of the product on path $p$, then the following Eq. (\ref{supplyFlow}) must be satisfied:

\begin{equation}\label{supplyFlow}
\sum\limits_{p \in {P_{{w_k}}}} {{x_p}}  = {d_{{w_k}}},\quad k = 1, \cdots ,{n_R}.
\end{equation}

Let $f_a$ represent the flow on link $a$, then the following conservation flow must be met:
\begin{equation}\label{conservationFlow}
{f_a} = \sum\limits_{p \in P} {{x_p}{\delta _{ap}}} ,\quad \forall a \in L.
\end{equation}
Eq. (\ref{conservationFlow}) means that the inflow must be equal to the outflow on link $a$.

These flows can be grouped into the vector $f$. The flow on each link must be a nonnegative number, i.e.
the following Eq. (\ref{constraint}) must be satisfied:
\begin{equation}\label{constraint}
{x_p} \ge 0,\quad \forall a \in L.
\end{equation}

Suppose the maximum capacity on link $a$ is expressed by $u_a, \forall a \in L$. It is required that the actual flow on link $a$ cannot exceed the maximum capacity on this link:
\begin{equation}\label{maximumFlow}
\begin{array}{l}
 {f_a} \le {u_a},\quad \forall a \in L, \\
 0 \le {u_a},\quad \forall a \in L. \\
 \end{array}
\end{equation}

The total cost on each link, for the simplicity, they can be represented as a function of the flow of the product on all the links \cite{nagurney2006supply,nagurney2009system,nagurney2002supply,nagurney2010environmental}:
\begin{equation}\label{constraint}
{\widehat c_a} = {\widehat c_a}\left( f \right),\quad \forall a \in L.
\end{equation}

The total investment cost of adding capacity $u_a$ on link $a$ can be expressed as follows:
\begin{equation}\label{capacityConstraint}
{\widehat\pi _a} = {\widehat\pi _a}\left( {{u_a}} \right),\quad \forall a \in L.
\end{equation}

Summarily, the supply chain network design optimization problem is to  satisfy the demand of each retail outlet
and  minimize the total cost, including the total cost of operating the various links and the capacity investments:
\begin{equation}\label{objective}
Minimize\;\;\sum\limits_{a \in L} {{{\widehat c}_a}\left( f \right)}  + \sum\limits_{a \in L} {{{\widehat\pi }_a}\left( {{u_a}} \right)}
\end{equation}
subject to constraints (\ref{supplyFlow})-(\ref{maximumFlow}).

\subsection{Physarum polycephalum}

{\em Physarum Polycephalum} is a large, single-celled amoeboid organism forming a dynamic tubular network connecting the discovered food sources during foraging. The mechanism of tube formation can be
described as follows. Tubes thicken in a given direction when shuttle streaming of the protoplasm persists in that direction for a certain time.  There is a positive feedback between flux and tube thickness, as the conductance of the sol is greater in a thicker channel. With this mechanism in mind, a mathematical model illustrating the shortest path finding has been constructed \cite{tero2007mathematical}.

Suppose the shape of the network formed by the \textit{Physarum} is represented by a graph, in
which a plasmodial tube refers to an edge of the graph and a junction
between tubes refers to a node. Two special nodes labeled as $N_1$, $N_2$ act as the starting node and ending node respectively. The other nodes are labeled as $N_3,N_4,N_5,N_6$ etc. The edge between nodes $N_i$ and $N_j$ is
$M_{ij}$. The parameter $Q_{ij}$ denotes the flux through tube $M_{ij}$ from node $N_i$ to $N_j$. Assume the flow along the tube is approximated by Poiseuille flow. Then the flux ${Q_{ij}}$ can be expressed as:
\begin{equation}\label{flux}
{Q_{ij}} = \frac{{{D_{ij}}}}{{{L_{ij}}}}({p_i} - {p_j})
\end{equation}
where ${p_i}$ is a pressure at a node ${N_i}$,
${D_{ij}}$ is a conductivity of a tube ${M_{ij}}$, and $L_{ij}$ is its length.

By assuming that the inflow and outflow must be balanced, we have:
\begin{equation}\label{sum}
\sum {{Q_{ij}}}  = 0\quad (j \ne 1,2)
\end{equation}

  For the source node ${N_1}$ and the sink node ${N_2}$ the
following holds:
\begin{equation}\label{soucesink}
\sum\limits_i {{Q_{i1}}}  + {I_0} = 0
\end{equation}
\begin{equation}\label{soucesink1}
\sum\limits_i {{Q_{i2}}} -{I_0} = 0
\end{equation}
where ${I_0}$ is the flux flowing from the source node and ${I_0}$ is a
constant value here.

  In order to describe such an adaptation of tubular thickness we
assume that the conductivity ${D_{ij}}$ changes over time
according to the flux ${Q_{ij}}$. An evolution of ${D_{ij}(t)}$ can be
described by the following equation:
\begin{equation}\label{changeovertime}
\frac{d}{{dt}}{D_{ij}} = f(|{Q_{ij}}|) - \gamma{D_{ij}}
\end{equation}
where $r$ is a decay rate of the tube. The equation implies that a
conductivity becomes nil if there is no flux along the edge. The conductivity increases
with the flux. The $f$ is monotonically increasing continuous function
satisfying $f(0) =0$.

  Then the network Poisson equation for the pressure can be
obtained from the Eq. (\ref{flux}-\ref{soucesink1}) as follows:
\begin{equation}\label{getresult}
\sum\limits_i {\frac{{{D_{ij}}}}{{{L_{ij}}}}({p_i} - {p_j})}  =
\left\{ {\begin{array}{*{20}{c}}
   { + 1} & {for} & {i = 1,}  \\
   { - 1} & {for} & {j = 2,}  \\
   0 & {otherwise.} & {}  \\
\end{array}} \right.
\end{equation}

  By setting ${p_2}$=0 as a basic pressure level, all ${p_i}$
can be determined by solving Eq. (\ref{getresult}) and ${Q_{ij}}$
can also be obtained.

In this paper, $f(Q) = |Q|$ is used because $f\left( {\left| {{Q_{ij}}} \right|} \right) = \left| Q \right|,\gamma  = 1$, the \emph{Physarum}
converges to the shortest path with a high rate \cite{tero2007mathematical}. With the flux calculated, the conductivity can be derived, where
Eq. (\ref{eql6}) is used instead of Eq.
(\ref{changeovertime}), adopting the functional form $f(Q) = |Q|$.
\begin{equation}\label{eql6}
\frac{{D_{ij}^{n + 1} - D_{ij}^n}}{{\delta t}} = |Q| - D_{ij}^{n +
1}
\end{equation}

\section{Proposed Method}

In this section, we  employ the \emph{Physarum} model to solve the supply chain network design problem. Generally speaking, there are two sub-problems to address:
\begin{itemize}
\item  In the shortest path finding model, there is only one source and one destination in the network while there are more than one retails in the supply chain network design problem.
\item The \emph{Physarum} model should be modified to satisfy the capacity constraint on each link.
\end{itemize}

\subsection{One-Source Multi-Sink's Physarum Model}

In the original \emph{Physarum} model \cite{tero2007mathematical}, there is only one source node and one ending node. In the supply chain network design problem, as shown in Fig. \ref{Topology}, there are $R_{n_R}$ retail outlets. From left to right, from top to bottom, we can number the nodes shown in Fig. \ref{Topology}. As a result, the following Eq. (\ref{supplyChain}) is formulated to replace the  Eq. (\ref{getresult}).

\begin{equation}\label{supplyChain}
\sum\limits_{i} { {\frac{{{D_{ij}}}}{{{L_{ij}}}}} \left( {{p_i} - {p_j}} \right)}  = \left\{ \begin{array}{l}
  + \sum\limits_{i = 1}^{n_R} {{d_{{R_i}}}} \;\;\;\;\;for\;j = 1 \\
  - d_{R_j} \;\;\;\;\;for\;j = {R_1},{R_2}, \cdots {R_{{n_R}}} \\
 \end{array} \right.
\end{equation}
where $j=1$ means that $\sum\limits_{i = 1}^{n_R} {{d_{{R_i}}}}$ units of goods are distributed from the firm to the other manufacturing facilities, $j =R_{{n_R}}$ denotes $n_R$ retail outlet needs $d_{R_{j}}$ units of goods.

In the original \emph{Physarum} model, the length associated with each link is fixed. In the supply chain network design problem, the cost on each link, be it a production link, a shipment link, or a storage link, is comprised of
 the cost of the flow on each link ${\widehat c_a}\left( f \right)$  and the investment cost ${\widehat\pi _a}\left( {{u_a}} \right)$. Practically, the flow on each link should be equal to its capacity. Assume there is flow $f_{ij}$ passing through the link $(i,j)$, we take the following measure to convert the two costs into one.

\begin{equation}\label{updatedCost}
{L_{ij}} = {{\hat c}_{ij}}\left( f_{ij} \right) + {{\hat \pi }_{ij}}\left( f_{ij} \right)
\end{equation}

 As a result, Eq. (\ref{updatedCost}) will be updated as below:
 \begin{equation}\label{supplyFlow1}
\sum\limits_i {\frac{{{D_{ij}}}}{{{{\hat c}_a}\left( {{f_{ij}}} \right) + {{\hat \pi }_a}\left( {{f_{ij}}} \right)}}\left( {{p_i} - {p_j}} \right)}  = \left\{ \begin{array}{l}
  + \sum\limits_{i = 1}^{{n_R}} {{d_{{R_i}}}} \quad for\;j = 1, \\
  - {d_{{R_j}}}\quad \quad for\;j = {R_1},{R_2}, \cdots {R_{{n_R}}}, \\
 0\quad \quad \quad otherwise \\
 \end{array} \right.
\end{equation}
where $f_{ij}$ represents the flow on the link starting from node $i$ to node $j$.

 In the \emph{Physarum} model, it is necessary for us to initialize the related parameters, including the link length $L$, the conductivity $D$, and the pressure $Q$ at each node. In the supply chain network design model, the capacity and the flow are the unknown design variables. If we don't know the specific flow on each link, we cannot determine its cost, which further leads to the initialization failure of \emph{Physarum} algorithm. To prevent this failure  we initialize the length on each link as a very small value ranging from 0.01 to 0.0001.

\subsection{Physarum-Inspired Model Satisfying the Imposed Capacity on Each Link}

To satisfy the imposed capacity on each link, we bring in a new parameter called Capacity Ratio (CR).
Let $f_{ij}$ be a flow on link $(i,j)$ is  $f_{ij}$ then the imposed capacity on  link $(i,j)$
is represented as $u_{ij}$.  The CR associated with this link is defined as follows:
\begin{equation}
C{R_{ij}} = \frac{{{f_{ij}}}}{{{u_{ij}}}}
\end{equation}

By defining this parameter, we aim at updating the cost on link $(i,j)$ as below:
\begin{equation}\label{CRupdate}
{L_{ij}} = C{R_{ij}} * \left( {{{\hat c}_a}\left( {{f_{ij}}} \right) + {{\hat \pi }_a}\left( {{f_{ij}}} \right)} \right) = \frac{{{f_{ij}}}}{{{u_{ij}}}} * \left( {{{\hat c}_a}\left( {{f_{ij}}} \right) + {{\hat \pi }_a}\left( {{f_{ij}}} \right)} \right)
\end{equation}

In this process, the parameter CR has an important role from two aspects. On the one hand, if the flow $f_{ij}$ exceeds its capacity $u_{ij}$, then $CR_{ij}$ will be bigger than 1, which leads to the increase of $L_{ij}$. Otherwise, $CR_{ij}$ will be less than 1, which further results in the decrease of $L_{ij}$.

According to the mechanism lying in the \emph{Physarum} model, every time the links' length changes, the flow along each link is  reallocated adaptively. Bonifaci et al. \cite{bonifaci2012physarum} has proved that the mass of the mould eventually converges to the shortest path.  Based on this rule, after a series of reallocation, we can reach the equilibrium state in the supply chain network design problem and the optimal supply chain network can be obtained.

\subsection{General Flow of Physarum Model}

The main flow of \emph{Physarum} model is presented in Algorithm \ref{algSupply}. The firm and the retail outlets are treated as the starting node and the ending node, respectively.

First of all, the conductivity of each tube (Dij) is initiated with random value
between $0$ and $1$ and other variables are assigned with $0$, including flux through
each tube ($Q_{ij}$), pressure at each node ($P_i$). 

Secondly, we can obtain the pressure associated with each node using Eq. (\ref{supplyFlow1}). Besides, the flux passing through each link and the conductivity in the next iteration can be recorded. Thirdly, we will update the cost on each link using Eq. (\ref{updatedCost}). In order to satisfy the capacity constraint, Eq. (\ref{CRupdate}) is imposed.

There are several possible solutions to decide when to stop execution of
Algorithm 1, such as the maximum number of iterations is arrived, conductivity of each tube converges to 0 or 1, flux through each tube remains unchanged, etc. The algorithm described in present paper halts when
$\sum\limits_{i = 1}^N {\sum\limits_{j = 1}^N {\left| {D_{ij}^{n } - D_{ij}^{n-1}} \right|} }  \le \delta$, where $\delta $ is a threshold value.

\linespread{1.1}
\begin{algorithm}[!ht]\scriptsize
\caption{\emph{Physarum}-Inspired Model for the Optimal Supply Chain Network Design (L,1,N,R)}\label{algSupply}
\begin{algorithmic}
\STATE // $N$ is  the size of the network;
\STATE // $L_{ij}$ is the link connecting node $i$ with node $j$;\\
\STATE // $1$ is the starting node while $R$ is the set of retail outlets; \\
${D_{ij}} \leftarrow \left( {0,1} \right]\;\left( {\forall i,j = 1,2, \ldots ,N} \right)$; \\
${Q_{ij}} \leftarrow 0\;\left( {\forall i,j = 1,2, \ldots ,N} \right)$; \\
${p_i} \leftarrow 0\;\left( {\forall i = 1,2, \ldots ,N} \right)$; \\
${L_{ij}} \leftarrow 0.001\left( {\forall i, j = 1,2, \ldots ,N} \right)$;\\
$count \leftarrow 1$ ;  \\
\REPEAT
   \STATE Calculate the pressure associated with each node according to the following Eq. (\ref{supplyFlow1})

  \[\sum\limits_i {\frac{{{D_{ij}}}}{{{{\hat c}_a}\left( {{f_{ij}}} \right) + {{\hat \pi }_a}\left( {{f_{ij}}} \right)}}\left( {{p_i} - {p_j}} \right)}  = \left\{ \begin{array}{l}
  + \sum\limits_{i = 1}^{{n_R}} {{d_{{R_i}}}} \quad for\;j = 1, \\
  - {d_{{R_j}}}\quad \quad for\;j = {R_1},{R_2}, \cdots {R_{{n_R}}}, \\
 0\quad \quad \quad otherwise \\
 \end{array} \right.\]

   \STATE ${Q_{ij}} \leftarrow {{{D_{ij}} \times \left( {{p_i} - {p_j}} \right)} \mathord{\left/
 {\vphantom {{{D_{ij}} \times \left( {{p_i} - {p_j}} \right)} {{L_{ij}}}}} \right.
 \kern-\nulldelimiterspace} {{L_{ij}}}}$ // Using Eq. (\ref{flux});

  \STATE ${D_{ij}} \leftarrow  {Q_{ij}}  + {D_{ij}}$  // Using Eq. (\ref{eql6})
  \STATE \bf{Update the cost on each link;}

  \FOR {$i=1:N$}
     \FOR {$j=1:N$}
       \IF {${Q_{ij}} \ne 0$}
       \STATE  $C{R_{ij}} = \frac{{\left| {{Q_{ij}}} \right|}}{{{u_{ij}}}};$
       \STATE  ${L_{ij}} = {L_{ij}} + {{\hat c}_{ij}}\left( {\left| {{Q_{ij}}} \right|} \right) + {{\hat \pi }_{ij}}\left( {\left| {{Q_{ij}}} \right|} \right);$
       \STATE ${L_{ij}} = {L_{ij}} * C{R_{ij}};$
       \STATE ${L_{ji}} = {L_{ji}};$
       \ENDIF
      \ENDFOR
  \ENDFOR

  \STATE $count \leftarrow count + 1$
\UNTIL{a termination criterion is met}
\end{algorithmic}
\end{algorithm}

\section{Numerical Examples}

In this section we demonstrate efficiency of the algorithm in numerical examples.
The supply chain network topology for all the examples is shown in Fig. \ref{example1},
and  $\delta $ is $0.0001$. In addition, we initialize the link length as $0.001$.

\begin{figure}[!ht]
\centering
\includegraphics[width=2in]{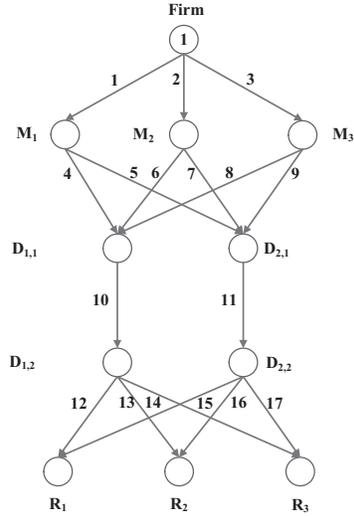}
\caption{The supply chain network topology for all the examples \cite{nagurney2010optimal}}
\label{AllExamples}
\end{figure}

\begin{figure}[!ht]
\centering
\includegraphics[width=5.6in]{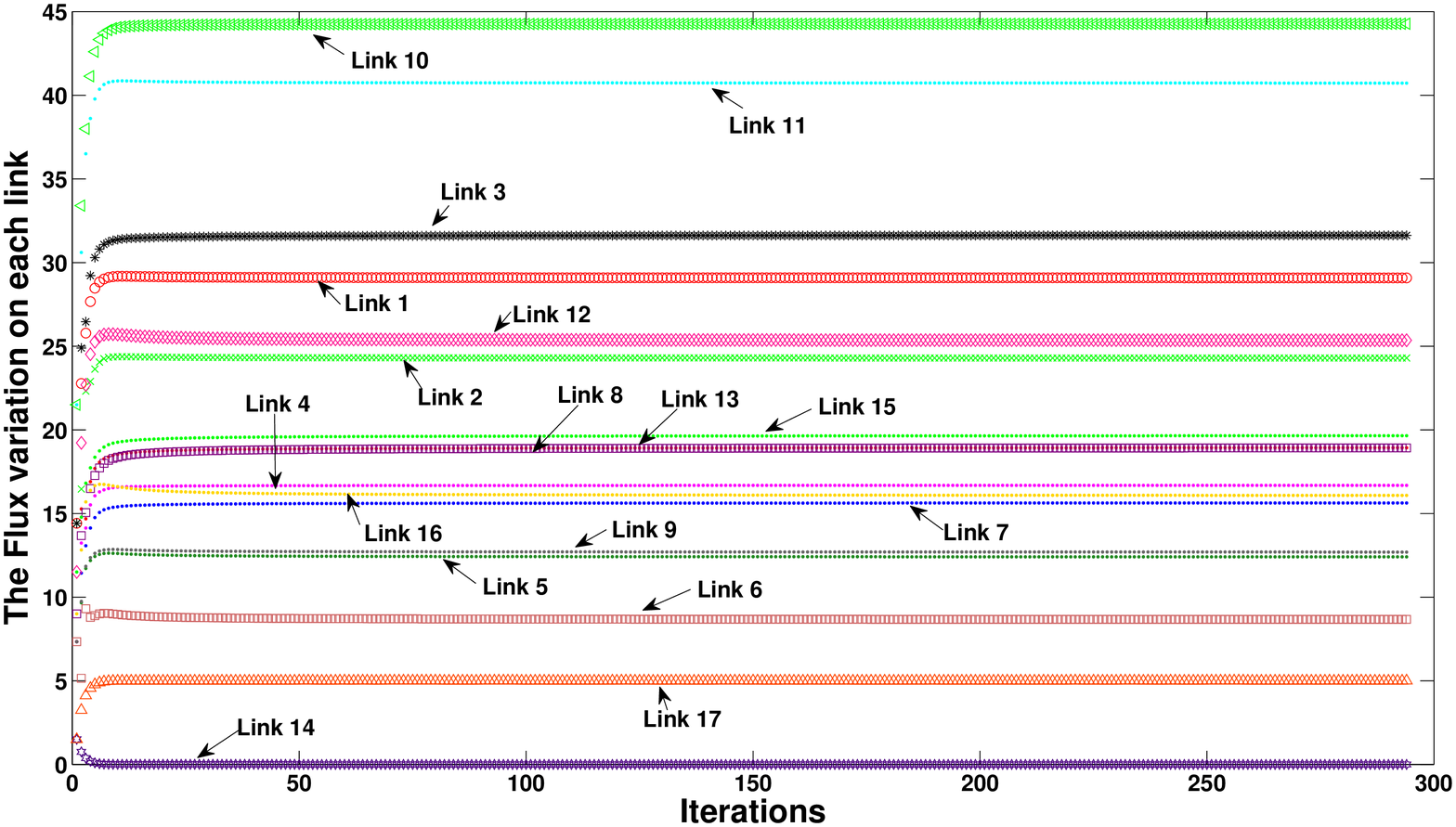}
\caption{The flux variation associated with each link during the iterative process in Example \ref{example1}}
\label{example1Flux}
\end{figure}

\linespread{1}
\begin{table}[!ht]\scriptsize
  \centering
  \caption{Total cost functions and solution in Example \ref{example1}. \label{supplyExample1} Adopted from Ref. \cite{nagurney2010optimal}}
  \begin{threeparttable}
    \begin{tabular}{lllll}
    \addlinespace
    \toprule
            Link $a$ &     ${\widehat c_a(f)}$   &  ${\widehat\pi _a}\left( {{u_a}} \right)$ & $f_a^*$ & $\lambda _a^*$\\
    \midrule
     1 & $f_1^2+2f_2$             &    $0.5u_1^2+u_1$                & 29.08 & 29.08   \\
     2 & $0.5f_2^2+f_2$           & $2.5u_2^2+u_2$                   & 24.29 & 24.29   \\
     3 & $0.5f_3^2+f_2$           & $u_3^2+2u_3$                     & 31.63 & 31.63   \\
     4 & $1.5f_4^2+2f_4$          & $u_4^2+u_4$                      & 16.68 & 16.68  \\
     5 & $f_5^2+3f_5$             & $2.5u_5^2+2u_5$                   & 12.40 & 12.40    \\
     6 & $f_6^2+2f_5$             & $0.5u_6^2+u_6$                   & 8.65  & 8.65    \\
     7 & $0.5f_7^2+2f_7$          & $0.5u_7^2+u_7$                   & 15.64 & 15.64    \\
     8 & $0.5f_8^2+2f_8$          & $1.5u_8^2+u_8$                   & 18.94 & 18.94   \\
     9 & $f_9^2+5f_9$             & $2u_9^2+3u_9$                    & 12.69 & 12.69    \\
     10 & $0.5f_{10}^2+2f_{10}$   &  $u_{10}^2+5u_{10}$              & 44.28 & 44.28    \\
     11 & $f_{11}^2+f_{11}$       & $0.5u_{11}^2+3u_{11}$            & 40.72 & 40.72    \\
     12 & $0.5f_{12}^2+2f_{12}$   & $0.5u_{12}^2+u_{12}$             & 25.34 & 25.34    \\
     13 & $0.5f_{13}^2+5f_{13}$   & $0.5u_{13}^2+u_{13}$             & 18.94 & 18.94   \\
     14 & $f_{14}^2+7f_{14}$      & $2u_{14}^2+5u_{14}$              & 0.00  & 0.00     \\
     15 & $f_{15}^2+2f_{15}$      & $0.5u_{15}^2+u_{15}$             & 19.66 & 19.66     \\
     16 & $0.5f_{16}^2+3f_{16}$   & $u_{16}^2+u_{16}$                & 16.06 & 16.06    \\
     17 & $0.5f_{17}^2+2f_{17}$   & $0.5u_{17}^2+u_{17}$             &  5.00 & 5.00  \\
    \bottomrule
    \end{tabular}
    \end{threeparttable}
\end{table}%

\begin{example}\label{example1}

In this example, the demands for each retail outlet is $d_{R_1}=45, d_{R_3}=35, d_{R_3}=5$, respectively. The cost of the flow on each link ${\widehat c_a}\left( f \right)$  and the investment cost ${\widehat\pi _a}\left( {{u_a}} \right)$ are shown in Table \ref{supplyExample1}; the costs are continuous-value functions.

Based on the proposed method, Fig. \ref{example1Flux} illustrates the flux variation during the iterative process.
The flux on each link gets stable with the increase of iterations. The flux of link $14$ corresponding to the shipment link connecting the first distribution center with the retailer $R_3$ gradually decreases to $0$ and it should be removed from the supply chain network design. Besides, the link $17$'s flux increases to $5$ step by step. Table \ref{supplyExample1} represents the solution. This solution corresponds to the optimal supply chain network design topology as shown
in Fig. \ref{optimalTopology}.

\begin{figure}[!ht]
\centering
\includegraphics[width=2in]{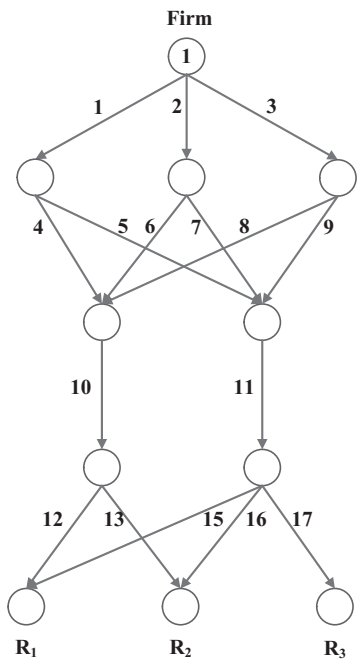}
\caption{Optimal supply chain network topology in Example \ref{example1}}
\label{optimalTopology}
\end{figure}

As for the objective of this problem as shown in Eq. (\ref{objective}), we can obtain the minimal cost. The result is 16125.65, which is in consistent with that of Ref. \cite{nagurney2010optimal}.
\end{example}

\begin{example}\label{example2}
The data in Example \ref{example2} has the same total cost as in Example \ref{example1} except that we use linear term to express the total cost associated with the first distribution center (For specific details, please see data for link 10 in Table \ref{supplyExample2}).

\begin{table}[!ht]\scriptsize
  \centering
  \caption{Total cost functions and solution for Example \ref{example2}. Adopted from Ref. \cite{nagurney2010optimal} \label{supplyExample2}}
  \begin{threeparttable}
    \begin{tabular}{lllll}
    \addlinespace
    \toprule
            Link a &     ${\widehat c_a(f)}$   &  ${\widehat\pi _a}\left( {{u_a}} \right)$ & $f_a^*$ & $\lambda _a^*$\\
    \midrule
     1 & $f_1^2+2f_2$             &    $0.5u_1^2+u_1$                & 29.28 & 29.28   \\
     2 & $0.5f_2^2+f_2$           & $2.5u_2^2+u_2$                   & 23.78 & 23.78   \\
     3 & $0.5f_3^2+f_2$           & $u_3^2+2u_3$                     & 31.93 & 31.93   \\
     4 & $1.5f_4^2+2f_4$          & $u_4^2+u_4$                      & 19.01 & 19.01 \\
     5 & $f_5^2+3f_5$             & $2.5u_5^2+2u_5$                  & 10.28 & 10.28    \\
     6 & $f_6^2+2f_5$             & $0.5u_6^2+u_6$                   & 13.73 & 13.73  \\
     7 & $0.5f_7^2+2f_7$          & $0.5u_7^2+u_7$                   & 10.05 & 10.05   \\
     8 & $0.5f_8^2+2f_8$          & $1.5u_8^2+u_8$                   & 21.77 & 21.77  \\
     9 & $f_9^2+5f_9$             & $2u_9^2+3u_9$                    & 10.17 & 10.17    \\
     10 & $0.5f_{10}^2+2f_{10}$   & $5u_{10}$                        & 54.50 & 54.50   \\
     11 & $f_{11}^2+f_{11}$       & $0.5u_{11}^2+3u_{11}$            & 30.50 & 30.50   \\
     12 & $0.5f_{12}^2+2f_{12}$   & $0.5u_{12}^2+u_{12}$             & 29.58 & 29.58    \\
     13 & $0.5f_{13}^2+5f_{13}$   & $0.5u_{13}^2+u_{13}$             & 23.18 & 23.18   \\
     14 & $f_{14}^2+7f_{14}$      & $2u_{14}^2+5u_{14}$              & 1.74  & 1.74    \\
     15 & $f_{15}^2+2f_{15}$      & $0.5u_{15}^2+u_{15}$             & 15.42 & 15.42   \\
     16 & $0.5f_{16}^2+3f_{16}$   & $u_{16}^2+u_{16}$                & 11.82 & 11.82  \\
     17 & $0.5f_{17}^2+2f_{17}$   & $0.5u_{17}^2+u_{17}$             & 3.26  & 3.26 \\
    \bottomrule
    \end{tabular}
    \end{threeparttable}
\end{table}%

Figure \ref{example2Flux} shows a changing trend of the flux associated with each link during the iterative process. The complete solution for this problem is shown in Table \ref{supplyExample2}. Tthe flow on link 14 now has positive capacity and positive product flow in contrast to  the data shown in Table \ref{supplyExample1}. As a matter of fact, the flow of all the links from the first distribution center to the retail outlets has increased, comparing to the values in Example \ref{example1}. Namely, the capacity and product flow on links 12, 13, and 14 is bigger than that of Example \ref{example1}. On the contrary, the product flow on the links connecting the second distribution center with the retail outlets decrease. For example, the solution values on links 15, 16, 17 in Example \ref{example2} are less when compared with that of Example \ref{example1}. Based on the solution shown in Table \ref{supplyExample2}, the final optimal supply chain network for this example can be formulated as shown in Fig. \ref{AllExamples}.

\begin{figure}[!ht]
\centering
\includegraphics[width=5.6in]{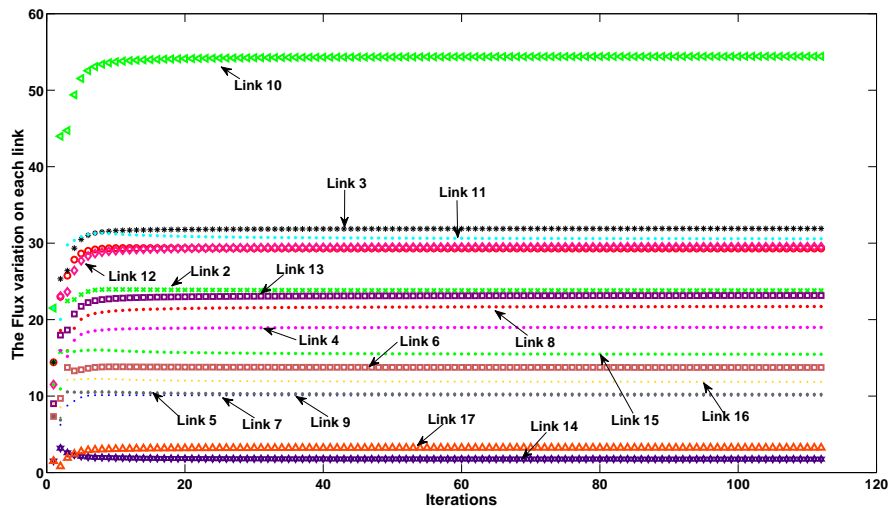}
\caption{The flux variation associated with each link during the iterative process in Example \ref{example2}}
\label{example2Flux}
\end{figure}

\end{example}

\begin{example}\label{example3}
Example \ref{example3} has the same data as Example \ref{example2} except that we use linear terms to replace  nonlinear functions representing the total costs associated with the capacity investments on the first and second manufacturing plants. For instance, the cost associated with the capacity investments on the first link is represented by $u_1$ instead of $0.5u_1^2+u_1$.

\begin{table}[!ht]\scriptsize
  \centering
  \caption{Total cost functions and solution for Example \ref{example3} \label{supplyExample3}. Adopted from Ref. \cite{nagurney2010optimal}}
  \begin{threeparttable}
    \begin{tabular}{lllll}
    \addlinespace
    \toprule
            Link a &     ${\widehat c_a(f)}$   &  ${\widehat\pi _a}\left( {{u_a}} \right)$ & $f_a^*$ & $\lambda _a^*$\\
    \midrule
     1 & $f_1^2+2f_2$             &    $u_1$                         & 20.91 & 20.91   \\
     2 & $0.5f_2^2+f_2$           & $u_2$                            & 45.18 & 45.18   \\
     3 & $0.5f_3^2+f_2$           & $u_3^2+2u_3$                     & 18.91 & 18.91   \\
     4 & $1.5f_4^2+2f_4$          & $u_4^2+u_4$                      & 14.74 & 14.74  \\
     5 & $f_5^2+3f_5$             & $2.5u_5^2+2u_5$                  & 6.16  & 6.16    \\
     6 & $f_6^2+2f_5$             & $0.5u_6^2+u_6$                   & 23.79 & 23.79 \\
     7 & $0.5f_7^2+2f_7$          & $0.5u_7^2+u_7$                   & 21.39 & 21.39  \\
     8 & $0.5f_8^2+2f_8$          & $1.5u_8^2+u_8$                   & 14.79 & 14.79  \\
     9 & $f_9^2+5f_9$             & $2u_9^2+3u_9$                    & 4.21  & 4.21    \\
     10 & $0.5f_{10}^2+2f_{10}$   & $5u_{10}$                        & 53.23 & 53.23  \\
     11 & $f_{11}^2+f_{11}$       & $0.5u_{11}^2+3u_{11}$            & 31.77 & 31.77   \\
     12 & $0.5f_{12}^2+2f_{12}$   & $0.5u_{12}^2+u_{12}$             & 29.10 & 29.10   \\
     13 & $0.5f_{13}^2+5f_{13}$   & $0.5u_{13}^2+u_{13}$             & 22.70 & 22.70   \\
     14 & $f_{14}^2+7f_{14}$      & $2u_{14}^2+5u_{14}$              & 1.44  & 1.44    \\
     15 & $f_{15}^2+2f_{15}$      & $0.5u_{15}^2+u_{15}$             & 15.90 & 15.90   \\
     16 & $0.5f_{16}^2+3f_{16}$   & $u_{16}^2+u_{16}$                & 12.30 & 12.30  \\
     17 & $0.5f_{17}^2+2f_{17}$   & $0.5u_{17}^2+u_{17}$             & 3.56  & 3.56 \\
    \bottomrule
    \end{tabular}
    \end{threeparttable}
\end{table}%

\begin{figure}[!ht]
\centering
\includegraphics[width=5.6in]{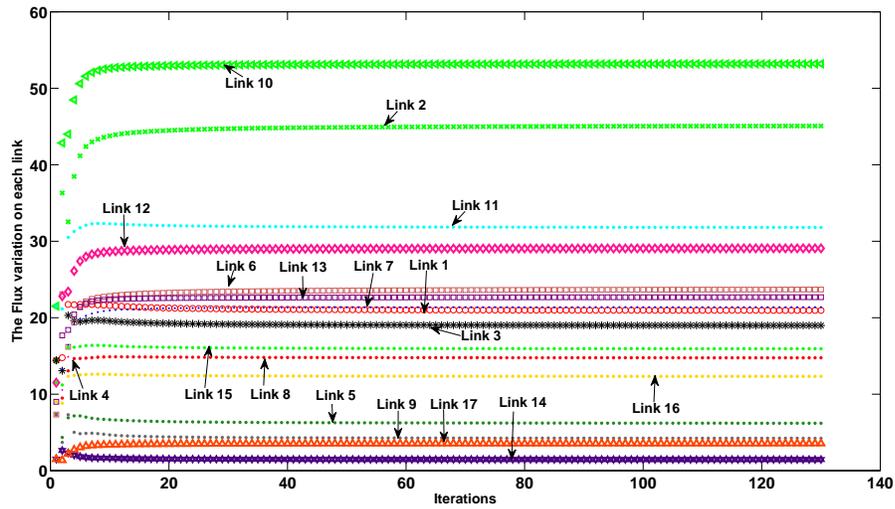}
\caption{The flux variation associated with each link during the iterative process in Example \ref{example3}}
\label{example3Flux}
\end{figure}

As can be seen from Fig. \ref{example3Flux}, it shows us the flux variation associated with each link during the iterative process. The solution for this problem is given in Table \ref{example3}. As for the objective function, our result is 10726.48, which is in accordance with that of Ref. \cite{nagurney2010optimal}. It can be noted that once the cost on each link changes, the proposed method can adaptively allocate the flow and the capacity investments.
\end{example}

\section{Conclusions}

We solve the supply chain network design problem using bio-inspired algorithm.  We propose a model for the supply chain network design allowing for the determination of the optimal levels of capacity and product flows in the supply activities, including manufacturing, distribution, and storage and subject to the satisfaction of retail outlets.
By employing principles of protoplasmic network growth and dynamical reconfiguration used by slime mould
 \emph{Physarum polycephalum} model, we solve the supply chain network design problem, no matter the cost associated with the capacity investments and the product flows is represented by linear or continuous functions.

Further research can focus on the following directions. First, we will adapt the method to the design
of supply chain network under complicated environment. For example, the supply chain network design problem with uncertain customer demands, the supply chain network redesign problem, and so on. Second, we will try to apply this method into other fields, such as the transportation network, mobile networks, and telecommunication networks etc.

\section*{Acknowledgement}
The work is partially supported Chongqing Natural Science Foundation, Grant No. CSCT, 2010BA2003,
National Natural Science Foundation of China, Grant No. 61174022,  National High Technology Research and Development Program of China (863 Program) (No.2013AA013801),
Doctor Funding of Southwest University Grant No. SWU110021.

\bibliographystyle{elsarticle-num-names}
\bibliography{ref}







\end{document}